\documentclass{article}

\usepackage{microtype}
\usepackage{graphicx}
\usepackage{subfigure}
\usepackage{newtxmath}
\usepackage{float}
\DeclareMathAlphabet{\mathpzc}{T1}{pzc}{m}{it}
\usepackage{booktabs} 

\usepackage{hyperref}

\usepackage[accepted]{icml2019}

\icmltitlerunning{Probabilistic Generative Models Harnessing Graph Neural Networks for Disease-Gene Prediction}

\begin{document}

\twocolumn[
\icmltitle{Towards Probabilistic Generative Models Harnessing Graph Neural Networks for Disease-Gene Prediction}




\begin{icmlauthorlist}
\icmlauthor{Vikash Singh}{to}
\icmlauthor{Pietro Lio'}{to}

\end{icmlauthorlist}

\icmlaffiliation{to}{Department of Computer Science and Technology, University of Cambridge, Cambridge, United Kingdom}

\icmlcorrespondingauthor{Vikash Singh}{vrs26@cam.ac.uk}
\icmlcorrespondingauthor{Pietro Lio'}{pl219@cam.ac.uk}

\icmlkeywords{Machine Learning, ICML}

\vskip 0.3in
]



\printAffiliationsAndNotice{} 

\begin{abstract}
Disease-gene prediction (DGP) refers to the computational challenge of predicting associations between genes and diseases. Effective solutions to the DGP problem have the potential to accelerate the therapeutic development pipeline at early stages via efficient prioritization of candidate genes for various diseases. In this work, we introduce the variational graph auto-encoder (VGAE) as a promising unsupervised approach for learning powerful latent embeddings in disease-gene networks that can be used for the DGP problem, the first approach using a generative model involving graph neural networks (GNNs). In addition to introducing the VGAE as a promising approach to the DGP problem, we further propose an extension (constrained-VGAE or C-VGAE) which adapts the learning algorithm for link prediction between two distinct node types in heterogeneous graphs. We evaluate and demonstrate the effectiveness of the VGAE on general link prediction in a disease-gene association network and the C-VGAE on disease-gene prediction in the same network, using popular random walk driven methods as baselines. While the methodology presented demonstrates potential solely based on utilizing the topology of a disease-gene association network, it can be further enhanced and explored through the integration of additional biological networks such as gene/protein interaction networks and additional biological features pertaining to the diseases and genes represented in the disease-gene association network. 

\end{abstract}

\section{Introduction}
\label{submission}
\subsection{Disease-Gene Networks}
In recent years, advances in technologies that allow for rapid acquisition and processing of biological data have increased tremendously, leading to an explosion of ``omics''  data, including genomics (pertaining to genetic sequences of DNA or RNA) and proteomics (pertaining to proteins, which are encoded through the expression of various genes) \cite{akula2009techniques} \cite{kodama2011sequence} \cite{gomez2014data}. Computational methods to process this vast set of data have also emerged as key drivers of growth in the field by uncovering associations between diseases and genes that can subsequently be validated through techniques in molecular biology \cite{quackenbush2001computational} \cite{cordell2009detecting}. The associations that have been validated form a disease-gene association network.

\subsection{The Disease-Gene Prediction Problem}
Disease-gene networks can be utilized by methodologies for prediction of the association between genes and diseases, tackling a computational challenge known as the disease-gene prediction (DGP) problem. There is significant interest in the DGP problem, as the genetic associations for many diseases are unknown and current techniques to validate associations between genes and diseases are both expensive and time-intensive, even with the adoption of next-generation sequencing technologies that have accelerated the process \cite{piro2012computational}. The network is characterized by an absence of explicit knowledge, allowing us to only categorize relationships as unknown or positive rather than directly categorize negative labels as with many other standard prediction problems \cite{madeddu2019network} \cite{venkatesan2009empirical}. From the perspective of therapeutics, many pharmaceutical solutions depend on specific protein or gene targets that have an established relationship with a particular disease. Once this relationship has been established, additional work is done to illuminate details of the biochemical pathways that govern a given association, which provides more insight on drug design. 

\subsection{Related Work}
Given the clinical relevance of the DGP problem, there has been significant interest in developing computational methodologies to tackle the problem over the past several years. The approaches can be categorized into three dominant classes: linkage methods, module-based methods, and diffusion methods \cite{barabasi2011network}. Very recently, representation learning has emerged as a class of approaches to tackle the problem. 

The foundation of linkage methods is built on the combination of candidate gene identification through network community analysis, knowledge of chromosomal locations of the candidate genes, and knowledge of certain disease loci (regions of the genome that are likely to be associated with a particular disease). In \cite{oti2006predicting}, candidate genes are identified through the analysis of a protein interaction network. Candidate genes are characterized by having an interaction with another gene (via corresponding proteins) known to have an association with a particular disease. From this perspective, a gene can qualify as a candidate if and only if it interacts with another gene that is known to have an association with the disease of interest, inherently limiting the scope of the search for candidate genes. 

Module-based methods are built on the concept that biological networks can be divided into modules or neighborhoods, which are roughly characterized by topological proximity. Candidate genes for a particular disease are assumed to belong to the same module as other genes that are known to be linked to that same disease. From the vantage point of pure network analysis, such a task can be framed as a community detection problem however, empirical analysis shows that genes/proteins that are associated with a particular disease do not always form dense subgraphs, although they may signify areas of functional similarity \cite{ghiassian2015disease}. 

Diffusion methods start with a set of seed genes which are known to be associated with a disease, however rather than computing a connectivity significance for each candidate gene in the network, random walk methods are used to find novel candidate genes \cite{vanunu2010associating} \cite{kohler2008walking} \cite{wu2008network} \cite{li2010genome} \cite{leiserson2015pan}.

In addition to the three traditional classes of approaches for tackling the DGP problem, representation learning approaches have recently emerged, which are focused on using graph-based methods to learn feature representations of nodes in relevant biological networks, often coupled with machine learning to identify candidate genes. A recent technique under this category is Random Watcher-Walker ($RW^2$), which jointly learns functional and connectivity patterns of proteins in a larger biological network containing protein-protein interactions \cite{madeddu2019network}. Another recent approach has established graph convolutional networks (GCNs) as a tool for tackling the DGP problem by combining information from the network topology and biologically relevant node features \cite{li2019pgcn}. 

\section{Methods}

\subsection{Dataset}
The disease-gene association network from the Stanford Biomedical Network Dataset Collection, which contains 7813 nodes in total (7294 genes, 519 diseases) is used to evaluate methods for the DGP problem. The disease-gene associations that comprise this network are collected from the Online Mendelian Inheritance in Man (OMIM) database, the Comparative Toxicogenomics Database (updated 2017), the DisGeNET dataset, a comprehensive platform integrating information on human disease-associated genes and variants, and MINER, a gigascale multimodal biological network developed by the Stanford SNAP group \cite{amberger2014omim} \cite{davis2016comparative} \cite{pinero2016disgenet}. 

\subsection{The Variational Graph Auto-Encoder (Standard)}

We describe the variational graph auto-encoder: \cite{kipf2016variational}

\textbf{Definitions}: 
We are given an undirected, unweighted graph $\mathcal{G}=(\mathcal{V},\mathcal{E})$ where $\mathcal{V}$ is the set of vertices and  $\mathcal{E}$ is the set of edges. We have $\mathcal{N}$ = $\mid \mathcal{V}\mid$ nodes. The adjacency matrix of $\mathcal{G}$ is A (with the assumption that each node shares an edge with itself, thus the diagonal elements of A are 1). We introduce D to be the degree matrix of $\mathcal{G}$. We introduce stochastic latent variables $z_{i}$ which can be characterized in an $N\times F$ matrix Z. Node features are characterized by a $N\times D$ matrix X. 

\vspace{5mm}

\noindent\textbf{Inference Model}:
We can frame the adjacency matrix reconstruction problem as a simple inference model, characterized by a two-layer GCN:
\begin{equation}
q(\textbf{Z} \mid \textbf{X}, \textbf{A}) = \prod_{i=1}^{N} q(\textbf{z}_i \mid \textbf{X}, \textbf{A}) 
\end{equation}

\text{  with   } $q(\textbf{z}_i)$ = $\mathpzc{N}(\textbf{z}_i \mid \boldsymbol{\mu}_i, \text{diag}(\boldsymbol{\sigma}_i^2))$

In this case, $\boldsymbol{\mu}$ = $GCN_{\mu}$(\textbf{X}, \textbf{A}) is the matrix of mean vectors $\boldsymbol{\mu}_i$; whereas $log(\boldsymbol{\sigma})$ = $GCN_{\sigma}$(\textbf{X}, \textbf{A}). We define the two-layer GCN as GCN(\textbf{X}, \textbf{A}) = $\widetilde{\textbf{A}}\text{ReLU}(\widetilde{\textbf{A}}\textbf{X}\textbf{W}_0)\textbf{W}_1$, with weight matrices $\textbf{W}_i$.

$GCN_{\mu}$(\textbf{X} \textbf{A}) and  $GCN_{\sigma}$(\textbf{X} \textbf{A}) share the parameters in the first layer defined by $\textbf{W}_0$. ReLU(n) = max(0,n) and  $\widetilde{\textbf{A}}= \textbf{D}^{-\frac{1}{2}}\textbf{A}\textbf{D}^{-\frac{1}{2}}$ is the symmetrically normalized adjacency matrix. 

\vspace{5mm}

\noindent\textbf{Generative Model}: We view the generative model as the reconstruction of the adjacency matrix from the latent variables obtained in the inference model outlined above. We use the inner product between the latent variable representations for two nodes to predict the presence of an association

\begin{equation}
p(\textbf{A} \mid \textbf{Z}) = \prod_{i=1}^N \prod_{j=1}^N  p(A_{ij} \mid \textbf{z}_i \text{, }\textbf{z}_j )
\end{equation}
\text{  with   } $p(A_{ij} =1 \mid \textbf{z}_i \text{, }\textbf{z}_j)$ = $\sigma(\textbf{z}_i^\top\textbf{z}_j)$

We define $A_{ij}$ to be the elements of \textbf{A} and $\sigma$ to be the sigmoid function. 

\vspace{5mm}

\noindent\textbf{Learning}: We optimize the variational lower bound $\mathcal{L}$ w.r.t. to the parameters $\textbf{W}_i$

\begin{equation}
\mathcal{L} = \mathbb{E}_{q(\textbf{Z} \mid \textbf{X}, \textbf{A})} [\text{log} p(\textbf{A} \mid \textbf{Z})] - \text{KL}[q(\textbf{Z} \mid \textbf{X}, \textbf{A}) \mid \mid p(\textbf{Z})]
\end{equation}

$KL[q\mid \mid p]$ represents the Kullback-Leibler divergence between q and p. We use a Gaussian prior p(\textbf{Z}) = $\prod_{i} p(\textbf{z}_i)$ = $\prod_{i} \mathpzc{N}(\textbf{z}_i \mid 0, \textbf{I})$. Gradient-based optimization such as batch or stochastic gradient descent is typically performed, and the reparametrization trick is utilized to allow for backpropagation on stochastic nodes of the architecture \cite{kingma2013auto}. For a graph in which the nodes do not have features, we replace \textbf{X} with the identity matrix in the GCN. 

\subsection{A Constrained VGAE for Specific Link Prediction in Heterogeneous Graphs}

The learning algorithm outlined in the standard VGAE does not focus on learning relationships specifically between distinct node types, optimizing for the reconstruction of the entire adjacency matrix given the latent representations that are stochastically derived from the GCN. While the optimization of the reconstruction of the entire adjacency matrix is intuitive in cases in which there is no desire to specifically learn the connectivity between two distinct classes of nodes, it is not intuitive in the case of disease-gene prediction in which we desire to specifically learn disease-gene relationships. Moreover, the method is not optimized for link prediction in heterogeneous graphs in which the desire is to learn associations between particular classes of nodes. In the case in which test edges are solely selected between the distinct node types of interest, the objective function to be optimized does not align with the prediction task. 

In order to adapt the standard VGAE for the specific link prediction task in a heterogeneous graph, the following extension to the methodology in introduced: 

\vspace{5mm}

We assume the same setup as the standard VGAE. In graph $\mathcal{G}$ however, the nodes in $\mathcal{V}$ can be split into $\mathcal{V}_0$ and $\mathcal{V}_1$, which represent the distinct node types (diseases and genes in the case of the DGP). Let the size of  $\mathcal{V}_0$ = $\mathcal{A}$ and the size of $\mathcal{V}_1$ = $\mathcal{B}$. Given that  $\mathcal{V}_0$ + $\mathcal{V}_1$ =  $\mathcal{V}$, we get that $\mathcal{A}$ + $\mathcal{B}$ = $\mathcal{N}$. We define $A^*$ to be a submatrix of the original adjacency matrix that solely represents edges between nodes in  $\mathcal{V}_0$ and nodes in  $\mathcal{V}_1$. $A^*$ has dimensions $\mathcal{A} \times \mathcal{B}$ and correspond to the Cartesian product between $\mathcal{V}_0$ and $\mathcal{V}_1$. Collectively, the elements of $A^*$ capture all the possible interactions between the two sets of nodes, and ignore any other relationships characterized in the original adjacency matrix A. 

Given this formulation of $A^*$, we adjust the learning paradigm of the standard VGAE as follows by altering the objective to be optimized 

\begin{equation}
\mathcal{L} = \mathbb{E}_{q(\textbf{Z} \mid \textbf{X}, \textbf{A})} [\text{log} p(\textbf{A}^* \mid \textbf{Z})] - \text{KL}[q(\textbf{Z} \mid \textbf{X}, \textbf{A}) \mid \mid p(\textbf{Z})]
\end{equation}

Rather than optimizing for the reconstruction of \textbf{A} given latent variables \textbf{Z}, we specifically optimize for the reconstruction of $\textbf{A}^*$ given $\textbf{Z}$, which causes the algorithm to focus on learning the relationships between nodes in $\mathcal{V}_0$  and $\mathcal{V}_1$. The adjacency matrices of heterogeneous graphs can be decomposed in submatrices representing the interactions of interest, giving rise to this adaptation to the optimization process. Due to the reduced reconstruction loss and the unchanged KL divergence term, it may be desirable to reweight the balance between the two terms in the loss function. 
Although the adjacency matrix reconstruction is reduced in scope, the structure of the GCN encoder does not change, allowing the embeddings to be derived from information collected throughout the entire network. 

\subsection{Evaluation}
Two sets of experiments are performed, with one involving the standard VGAE for general link prediction (not confined to solely predicting disease-gene associations), and the other solely focusing on disease-gene prediction, in which case the C-VGAE is used. Both DeepWalk and node2vec are demonstrated as baselines, due to their strong performance on benchmark link prediction tasks and similarity in methodology to many previous approaches for the DGP problem, utilizing random walks to collect relevant information from the network \cite{perozzi2014deepwalk} \cite{grover2016node2vec}. DeepWalk and node2vec algorithms are set to create embeddings with 128 dimensions, utilizing an inner product decoder with a sigmoid function to produce the final link predictions. For both DeepWalk and node2vec, the number of walks is 10, the length of each walk is 80, and the window size is 10. For node2vec, the p and q parameter values are set to 1. 

Both the VGAE and C-VGAE utilize a GCN with 200 nodes in the hidden layer and 20 nodes in the output layer, using dropout as a regularization method with a keep probability of 0.5. The Adam optimizer is used in both models with a learning rate of 0.05, and trained for 2 iterations. 80\% of the edges are used for training, with 10\% in a validation set for parameter tuning and 10\% in a test set to evaluate final metrics. We report \textit{area under the ROC curve} (AUC) and \textit{average precision} (AP) scores for each model on the test set. The mean and standard error for 10 runs of each experiment are displayed.  

\section{Results and Discussion}

\begin{table}[H]
\caption{Performance on General Link Prediction in Disease-Gene Association Network}
\label{sample-table}
\vskip 0.15in
\begin{center}
\begin{small}
\begin{sc}
\begin{tabular}{lcccr}
\toprule
Method & AUC & AP\\
\midrule
DeepWalk    & 79.5$\pm$ 7.30& 81.4$\pm$ 7.31 \\
Node2Vec & 79.6$\pm$ 5.40& 79.2$\pm$ 4.68\\
VGAE    & \textbf{84.4}$\pm$ 1.65& \textbf{86.4}$\pm$ 1.39 \\

\bottomrule
\end{tabular}
\end{sc}
\end{small}
\end{center}
\vskip -0.1in
\end{table}

\begin{table}[H]
\caption{Performance on Disease-Gene Prediction in Disease-Gene Association Network}
\label{sample-table}
\vskip 0.15in
\begin{center}
\begin{small}
\begin{sc}
\begin{tabular}{lcccr}
\toprule
Method & AUC & AP\\
\midrule
DeepWalk    & 78.5$\pm$ 2.25& 77.4$\pm$ 2.51 \\
Node2Vec & 86.0$\pm$ 1.98& 84.6$\pm$ 2.36\\
C-VGAE    & \textbf{90.8}$\pm$ 1.72& \textbf{91.3}$\pm$ 1.69 \\

\bottomrule
\end{tabular}
\end{sc}
\end{small}
\end{center}
\vskip -0.1in
\end{table}

In both sets of experiments as seen in Table 1 and 2, the VGAE and C-VGAE outperform DeepWalk and node2vec, two popular link prediction methods that are driven through random walks in the network, similar to many proposed diffusion methods for the DGP. These results demonstrate the potential for a generative graph neural network driven methodology via the VGAE to capture powerful latent structure that can be used for link prediction tasks in disease-gene networks. The C-VGAE implementation demonstrates promise as a modification of standard VGAEs for specific link prediction tasks in heterogeneous graphs, which are extremely common in real-world settings.

Since the space of latent embeddings for nodes in the network is stochastic, the VGAE can fundamentally be viewed as a generative model as new reconstructions of the adjacency matrix (corresponding to new link predictions) can be performed by resampling latent node representations. This generative nature can be utilized in the DGP problem to create and analyze multiple sets of predictions which provide a more comprehensive view of the interactions that may occur in the network. Obtaining multiple sets of predictions via resampling latent node representations may also lead to developing a more rigorous notion of confidence for each of the predicted associations.   

Compared to linkage methods for the DGP problem, this methodology harnessing graph neural networks is not limited by incomplete knowledge of disease-associated genomic loci.  Compared to module-based methods, candidate genes detected in a graph neural network approach are not confined to specific modules in the network, leading to more flexibility, especially since it has been shown that genes associated with a disease don't always form dense subgraphs. With the results and methodology presented as a baseline, future work involving integration with other datasets including gene/protein interaction networks, and gene/disease features looks promising, especially since the outlined approach has the flexibility to seamlessly integrate such features compared to other approaches. In addition to integration with other data sources, more experimentation on comparisons with existing approaches will be insightful. For computational methodologies focused on disease-gene prediction such as the one presented in this paper, integrating gene ontology (GO) analysis may additionally be a useful tool for the validation of predicted associations \cite{zheng2008goeast}.

\section{Conclusion}

In conclusion, we present a probabilistic generative approach involving graph neural networks for link prediction on disease-gene networks via the VGAE, and more specifically disease-gene prediction. We demonstrate its potential against popular link prediction methods which harness random walks on a recently published disease-gene association network. We further present the C-VGAE as an extension to the standard VGAE that demonstrates promise for specific link prediction tasks in heterogeneous graphs. With the C-VGAE, we present a foundation for adaptations of unsupervised graph neural network methods to specific types of link prediction tasks in heterogeneous graphs, which are prevalent in real-world link prediction problems. As graph neural network approaches continue to develop, it is clear their application in the domain of biological sciences will be critical. The work presented takes a step towards bridging the gap between graph neural network driven methods and computational challenges in biological sciences, while proposing a more generalizable extension to the VGAE for specific link prediction in heterogeneous graphs.

\

\bibliography{example_paper}

\begin{thebibliography}{25}
\providecommand{\natexlab}[1]{#1}
\providecommand{\url}[1]{\texttt{#1}}
\expandafter\ifx\csname urlstyle\endcsname\relax
  \providecommand{\doi}[1]{doi: #1}\else
  \providecommand{\doi}{doi: \begingroup \urlstyle{rm}\Url}\fi

\bibitem[Akula et~al.(2009)Akula, Miriyala, Thota, Rao, and
  Gedela]{akula2009techniques}
Akula, S.~P., Miriyala, R.~N., Thota, H., Rao, A.~A., and Gedela, S.
\newblock Techniques for integrating-omics data.
\newblock \emph{Bioinformation}, 3\penalty0 (6):\penalty0 284, 2009.

\bibitem[Amberger et~al.(2014)Amberger, Bocchini, Schiettecatte, Scott, and
  Hamosh]{amberger2014omim}
Amberger, J.~S., Bocchini, C.~A., Schiettecatte, F., Scott, A.~F., and Hamosh,
  A.
\newblock Omim. org: Online mendelian inheritance in man
  (omim{\textregistered}), an online catalog of human genes and genetic
  disorders.
\newblock \emph{Nucleic acids research}, 43\penalty0 (D1):\penalty0 D789--D798,
  2014.

\bibitem[Barab{\'a}si(2011)]{barabasi2011network}
Barab{\'a}si, A.-L.
\newblock The network takeover.
\newblock \emph{Nature Physics}, 8\penalty0 (1):\penalty0 14, 2011.

\bibitem[Cordell(2009)]{cordell2009detecting}
Cordell, H.~J.
\newblock Detecting gene--gene interactions that underlie human diseases.
\newblock \emph{Nature Reviews Genetics}, 10\penalty0 (6):\penalty0 392, 2009.

\bibitem[Davis et~al.(2016)Davis, Grondin, Johnson, Sciaky, King, McMorran,
  Wiegers, Wiegers, and Mattingly]{davis2016comparative}
Davis, A.~P., Grondin, C.~J., Johnson, R.~J., Sciaky, D., King, B.~L.,
  McMorran, R., Wiegers, J., Wiegers, T.~C., and Mattingly, C.~J.
\newblock The comparative toxicogenomics database: update 2017.
\newblock \emph{Nucleic acids research}, 45\penalty0 (D1):\penalty0 D972--D978,
  2016.

\bibitem[Ghiassian et~al.(2015)Ghiassian, Menche, and
  Barab{\'a}si]{ghiassian2015disease}
Ghiassian, S.~D., Menche, J., and Barab{\'a}si, A.-L.
\newblock A disease module detection (diamond) algorithm derived from a
  systematic analysis of connectivity patterns of disease proteins in the human
  interactome.
\newblock \emph{PLoS computational biology}, 11\penalty0 (4):\penalty0
  e1004120, 2015.

\bibitem[Gomez-Cabrero et~al.(2014)Gomez-Cabrero, Abugessaisa, Maier,
  Teschendorff, Merkenschlager, Gisel, Ballestar, Bongcam-Rudloff, Conesa, and
  Tegn{\'e}r]{gomez2014data}
Gomez-Cabrero, D., Abugessaisa, I., Maier, D., Teschendorff, A.,
  Merkenschlager, M., Gisel, A., Ballestar, E., Bongcam-Rudloff, E., Conesa,
  A., and Tegn{\'e}r, J.
\newblock Data integration in the era of omics: current and future challenges,
  2014.

\bibitem[Grover \& Leskovec(2016)Grover and Leskovec]{grover2016node2vec}
Grover, A. and Leskovec, J.
\newblock node2vec: Scalable feature learning for networks.
\newblock In \emph{Proceedings of KDD}, pp.\  855--864, 2016.

\bibitem[Kingma \& Welling(2013)Kingma and Welling]{kingma2013auto}
Kingma, D.~P. and Welling, M.
\newblock Auto-encoding variational bayes.
\newblock \emph{arXiv preprint arXiv:1312.6114}, 2013.

\bibitem[Kipf \& Welling(2016)Kipf and Welling]{kipf2016variational}
Kipf, T.~N. and Welling, M.
\newblock Variational graph auto-encoders.
\newblock \emph{arXiv preprint arXiv:1611.07308}, 2016.

\bibitem[Kodama et~al.(2011)Kodama, Shumway, and Leinonen]{kodama2011sequence}
Kodama, Y., Shumway, M., and Leinonen, R.
\newblock The sequence read archive: explosive growth of sequencing data.
\newblock \emph{Nucleic acids research}, 40\penalty0 (D1):\penalty0 D54--D56,
  2011.

\bibitem[K{\"o}hler et~al.(2008)K{\"o}hler, Bauer, Horn, and
  Robinson]{kohler2008walking}
K{\"o}hler, S., Bauer, S., Horn, D., and Robinson, P.~N.
\newblock Walking the interactome for prioritization of candidate disease
  genes.
\newblock \emph{The American Journal of Human Genetics}, 82\penalty0
  (4):\penalty0 949--958, 2008.

\bibitem[Leiserson et~al.(2015)Leiserson, Vandin, Wu, Dobson, Eldridge, Thomas,
  Papoutsaki, Kim, Niu, McLellan, et~al.]{leiserson2015pan}
Leiserson, M.~D., Vandin, F., Wu, H.-T., Dobson, J.~R., Eldridge, J.~V.,
  Thomas, J.~L., Papoutsaki, A., Kim, Y., Niu, B., McLellan, M., et~al.
\newblock Pan-cancer network analysis identifies combinations of rare somatic
  mutations across pathways and protein complexes.
\newblock \emph{Nature genetics}, 47\penalty0 (2):\penalty0 106, 2015.

\bibitem[Li \& Patra(2010)Li and Patra]{li2010genome}
Li, Y. and Patra, J.~C.
\newblock Genome-wide inferring gene--phenotype relationship by walking on the
  heterogeneous network.
\newblock \emph{Bioinformatics}, 26\penalty0 (9):\penalty0 1219--1224, 2010.

\bibitem[Li et~al.(2019)Li, Kuwahara, Yang, Song, and Gao]{li2019pgcn}
Li, Y., Kuwahara, H., Yang, P., Song, L., and Gao, X.
\newblock Pgcn: Disease gene prioritization by disease and gene embedding
  through graph convolutional neural networks.
\newblock \emph{bioRxiv}, pp.\  532226, 2019.

\bibitem[Madeddu et~al.(2019)Madeddu, Stilo, and Velardi]{madeddu2019network}
Madeddu, L., Stilo, G., and Velardi, P.
\newblock Network-based methods for disease-gene prediction.
\newblock \emph{arXiv preprint arXiv:1902.10117}, 2019.

\bibitem[Oti et~al.(2006)Oti, Snel, Huynen, and Brunner]{oti2006predicting}
Oti, M., Snel, B., Huynen, M.~A., and Brunner, H.~G.
\newblock Predicting disease genes using protein--protein interactions.
\newblock \emph{Journal of medical genetics}, 43\penalty0 (8):\penalty0
  691--698, 2006.

\bibitem[Perozzi et~al.(2014)Perozzi, Al-Rfou, and Skiena]{perozzi2014deepwalk}
Perozzi, B., Al-Rfou, R., and Skiena, S.
\newblock Deepwalk: Online learning of social representations.
\newblock In \emph{Proceedings of KDD}, pp.\  701--710, 2014.

\bibitem[Pi{\~n}ero et~al.(2016)Pi{\~n}ero, Bravo, Queralt-Rosinach,
  Guti{\'e}rrez-Sacrist{\'a}n, Deu-Pons, Centeno, Garc{\'\i}a-Garc{\'\i}a,
  Sanz, and Furlong]{pinero2016disgenet}
Pi{\~n}ero, J., Bravo, {\`A}., Queralt-Rosinach, N.,
  Guti{\'e}rrez-Sacrist{\'a}n, A., Deu-Pons, J., Centeno, E.,
  Garc{\'\i}a-Garc{\'\i}a, J., Sanz, F., and Furlong, L.~I.
\newblock Disgenet: a comprehensive platform integrating information on human
  disease-associated genes and variants.
\newblock \emph{Nucleic acids research}, pp.\  gkw943, 2016.

\bibitem[Piro \& Di~Cunto(2012)Piro and Di~Cunto]{piro2012computational}
Piro, R.~M. and Di~Cunto, F.
\newblock Computational approaches to disease-gene prediction: rationale,
  classification and successes.
\newblock \emph{The FEBS journal}, 279\penalty0 (5):\penalty0 678--696, 2012.

\bibitem[Quackenbush(2001)]{quackenbush2001computational}
Quackenbush, J.
\newblock Computational genetics: computational analysis of microarray data.
\newblock \emph{Nature reviews genetics}, 2\penalty0 (6):\penalty0 418, 2001.

\bibitem[Vanunu et~al.(2010)Vanunu, Magger, Ruppin, Shlomi, and
  Sharan]{vanunu2010associating}
Vanunu, O., Magger, O., Ruppin, E., Shlomi, T., and Sharan, R.
\newblock Associating genes and protein complexes with disease via network
  propagation.
\newblock \emph{PLoS computational biology}, 6\penalty0 (1):\penalty0 e1000641,
  2010.

\bibitem[Venkatesan et~al.(2009)Venkatesan, Rual, Vazquez, Stelzl, Lemmens,
  Hirozane-Kishikawa, Hao, Zenkner, Xin, Goh, et~al.]{venkatesan2009empirical}
Venkatesan, K., Rual, J.-F., Vazquez, A., Stelzl, U., Lemmens, I.,
  Hirozane-Kishikawa, T., Hao, T., Zenkner, M., Xin, X., Goh, K.-I., et~al.
\newblock An empirical framework for binary interactome mapping.
\newblock \emph{Nature methods}, 6\penalty0 (1):\penalty0 83, 2009.

\bibitem[Wu et~al.(2008)Wu, Jiang, Zhang, and Li]{wu2008network}
Wu, X., Jiang, R., Zhang, M.~Q., and Li, S.
\newblock Network-based global inference of human disease genes.
\newblock \emph{Molecular systems biology}, 4\penalty0 (1):\penalty0 189, 2008.

\bibitem[Zheng \& Wang(2008)Zheng and Wang]{zheng2008goeast}
Zheng, Q. and Wang, X.-J.
\newblock Goeast: a web-based software toolkit for gene ontology enrichment
  analysis.
\newblock \emph{Nucleic acids research}, 36\penalty0 (suppl\_2):\penalty0
  W358--W363, 2008.

\end{thebibliography}
\bibliographystyle{icml2019}



\end{document}